\documentclass[runningheads]{llncs}

\usepackage{graphicx}
\usepackage{booktabs}
\usepackage{CJK}
\usepackage{float}
\usepackage{times}
\usepackage{threeparttable}

\begin{document}
%
\title{Robust Learning for Text Classification with Multi-source Noise Simulation and Hard Example Mining}

\titlerunning{Robust Learning for Text Classification}
%
\author{Guowei Xu, Wenbiao Ding\thanks{Corresponding Author: Wenbiao Ding}, Weiping Fu, Zhongqin Wu, Zitao Liu}

\authorrunning{G. Xu et al.}
%
\institute{TAL Education Group, Beijing, China \\
\email{\{xuguowei, dingwenbiao, fuweiping1, wuzhongqin, liuzitao\}@tal.com}}

%
%
%
%
%
%
\maketitle              
\begin{abstract}
Many real-world applications involve the use of Optical Character Recognition (OCR) engines to transform handwritten images into transcripts on which downstream Natural Language Processing (NLP) models are applied. In this process, OCR engines may introduce errors and inputs to downstream NLP models become noisy. Despite that pre-trained models achieve state-of-the-art performance in many NLP benchmarks, we prove that they are not robust to noisy texts generated by real OCR engines. This greatly limits the application of NLP models in real-world scenarios. In order to improve model performance on noisy OCR transcripts, it is natural to train the NLP model on labelled noisy texts. However, in most cases there are only labelled clean texts. Since there is no handwritten pictures corresponding to the text, it is impossible to directly use the recognition model to obtain noisy labelled data. Human resources can be employed to copy texts and take pictures, but it is extremely expensive considering the size of data for model training. Consequently, we are interested in making NLP models intrinsically robust to OCR errors in a low resource manner. We propose a novel robust training framework which 1) employs simple but effective methods to directly simulate natural OCR noises from clean texts and 2) iteratively mines the hard examples from a large number of simulated samples for optimal performance. 3) To make our model learn noise-invariant representations, a stability loss is employed. Experiments on three real-world datasets show that the proposed framework boosts the robustness of pre-trained models by a large margin. We believe that this work can greatly promote the application of NLP models in actual scenarios, although the algorithm we use is simple and straightforward. We make our codes and three datasets publicly available\footnote{https://github.com/tal-ai/Robust-learning-MSSHEM}.
\end{abstract}
\section{Introduction}
\label{sec:intro}
With the help of deep learning models, significant advances have been made in different NLP tasks. In recent years, pre-trained models such as BERT \cite{Devlin2019BERTPO} and its variants achieved state-of-the-art performance in many NLP benchmarks. While human being can easily process noisy texts that contain typos, misspellings, and the complete omission of letters when reading \cite{rawlinson2007significance}, most NLP systems fail when processing corrupted or noisy texts \cite{belinkov2017synthetic}. It is not intuitive, however, if pre-trained NLP models are robust under noisy text setting.

There are several scenarios in which noise could be generated. The first type is user-generated noise. Typos and misspellings are the major ones and they are commonly introduced when users input texts through keyboards. Some other user-generated noise includes incorrect use of tense, singular and plural, etc. The second type of noise is machine-generated. A typical example is in the essay grading system \cite{valenti2003overview}. Students upload images of handwritten essays to the grader system in which OCR engines transform images to structured texts. In this process, noise is introduced in texts and it can make downstream NLP models fail. We argue that the distribution of user-generated errors is different from that of OCR errors. For example, people often mistype characters that are close to each other on the keyboards, or make grammatical mistakes such as incorrect tense, singular and plural. However, OCR is likely to misrecognize similar handwritten words such as ``dog" and ``dag", but it it unlikely to make mistakes that are common for humans.

There are many existing works \cite{sun2020adv,sun2019contextual} on how to improve model performance when there are user-generated noises in inputs. \cite{sun2020adv} studied the character distribution on the keyboard to simulate real user-generated texts for BERT. \cite{sun2019contextual} employed masked language models to denoise the input so that model performance on downstream task improves. Another existing line of work focuses on adversarial training, which refers to applying a small perturbation on the model input to craft an adversarial example, ideally imperceptible by humans, and causes the model to make an incorrect prediction \cite{goodfellow2014explaining}. It is believed that model trained on adversarial data is more robust than model trained on clean texts. However, adversarial attack focuses on the weakness in NLP models but does not consider the distribution of OCR errors, so the generated sample is not close to natural OCR transcripts, making adversarial training less effective in our problem.
 
Despite that NLP models are downstream of OCR engines in many real-world applications, there are few works on how to make NLP models intrinsically robust to natural OCR errors. In this paper, we discuss how the performance of pre-trained models degrades on natural OCR transcripts in text classification and how can we improve its robustness on the downstream task. We propose a novel robust learning framework that largely boosts the performance of pre-trained models when evaluated on both noise-free data and natural OCR transcripts in text classification task. We believe that this work can greatly promote the application of NLP models in actual noise scenarios, although the algorithm we use is simple and straightforward. Our contributions are:
\begin{itemize}

\item {We propose three simple but effective methods, rule-based, model-based and attack-based simulation, to generate natural OCR noises.}

\item{In order to combine the noise simulation methods, we propose a hard example mining algorithm so that the model focuses more on hard samples in each epoch of training. We define hard examples as those whose representations are quite different between noise-free inputs and noisy inputs. This ensures that the model learns more robust representations compared to naively treating all simulated samples equally.}

\item{We evaluate the framework on three real-world datasets and prove that the proposed framework outperforms existing robust training approaches by a large margin.}


\item{We make our code and data publicly available. To the best of our knowledge, we are the first to evaluate model robustness on OCR transcripts generated by real-world OCR engines.}
\end{itemize}

\section{Related Work}
\label{sec:related}
\subsection{Noise Reduction}
An existing approach to deal with noisy inputs is to introduce some denoising modules into the system. Grammatical Error Correction (GEC) systems have been widely used to address this problem. Simple rule-based and frequency-based spell-checker \cite{ndiaye2003spell} are limited to complex language systems. More recently, modern neural GEC systems are developed with the help of deep learning \cite{zhao2019improving,chollampatt2018neural}. Despite that neural GEC achieves SOTA performance, there are at least two problems with using GEC as a denoising module to alleviate the impact of OCR errors. Firstly, it requires a massive amount of parallel data, e.g., \cite{sang2003introduction} to train a neural GEC model, which is expensive to acquire in many scenarios. Secondly, GEC systems can only correct user-generated typos, misspellings and grammatical errors, but the distribution of these errors is quite different from that of OCR errors, making GEC limited as a denoiser. For example, people often mistype characters that are close to each other on the keyboards, or make grammatical mistakes such as tense, singular and plural. However, OCR is likely to misrecognize similar handwritten words such as ``dog" and ``dag", but it is unlikely to make mistakes that are common for humans. Another line of research focuses on how to use language models \cite{zhai2008statistical} as the denoising module. \cite{sun2019contextual} proposed to use masked language models in an off-the-shelf manner. Although this approach does not rely on massive amount of parallel data, it still oversimplifies the problem by not considering OCR error distributions. More importantly, we are interested in boosting intrinsic model robustness. In other words, if we directly feed noisy data into the classification model, it should be able to handle it without relying on extra denoising modules. However, both GEC and language model approaches are actually pre-processing modules, and they do not improve the intrinsic robustness of downstream NLP models. Therefore, we do not experiment on denoising modules in this paper.

\subsection{Adversarial Training}
Adversarial attack aims to break down neural models by adding imperceptible perturbations on the input. Adversarial training \cite{miyato2016adversarial,yasunaga2017robust} improves the robustness of neural networks by training models on adversarial samples. There are two types of adversarial attacks, the white-box attack \cite{Ebrahimi2018HotFlipWA} and the black-box attack \cite{alzantot-etal-2018-generating,zhao2017generating}. The former assumes access to the model parameters when generating adversarial samples while the latter can only observe model outputs given attacked samples. Recently, there are plenty of works on attacking NLP models. \cite{ribeiro2018semantically} found that NLP models often make different predictions for texts that are semantically similar, they summarized simple replacement rules from these semantically similar texts and re-trained NLP models by augmenting training data to address this problem. \cite{sun2020adv} proved that BERT is not robust to typos and misspellings and re-trained it with nature adversarial samples. Although it has been proved that adversarial training is effective to improve the robustness of neural networks, it searches for weak spots of neural networks but does not consider common OCR errors in data augmentation. Therefore, traditional adversarial training is limited in our problem.

\subsection{Training with Noisy Data}
Recent work has proved that training with noisy data can boost NLP model performance to some extent. \cite{belinkov2017synthetic} pointed out that a character-based CNN trained on noisy data can learn robust representations to handle multiple kinds of noise. \cite{karpukhin2019training} created noisy data using random character swaps, substitutions, insertions and deletions and improved model performance in machine translation under permuted inputs. \cite{namysl2020nat} simulated noisy texts using a confusion matrix and employed a stability loss when training models on both clean and noisy samples.

In this paper, our robust training framework follows the same idea to train models with both clean and noisy data. The differences are that our multi-source noise simulation can generate more natural OCR noises and using hard example mining algorithm together with stability loss can produce optimal performance.

\section{Problem}
\label{problem}
\subsection{Notation}
In order to distinguish noise-free texts, natural handwritten OCR transcripts and simulated OCR transcripts, we denote them by $\mathcal{X}$, $\mathcal{X^\prime}$ and $\mathcal{\widetilde{X}}$ respectively. Let $\mathcal{Y}$ denote the shared labels. 

\subsection{Text Classification}
Text classification is one of the most common NLP tasks and can be used to evaluate the performance of NLP models. Text classification is the assignment of documents to a fixed number of semantic categories. Each document can be in multiple or exactly one category or no category at all. More formally, let $\mathbf{x}=(\mathbf{w_0}, \mathbf{w_1}, \mathbf{w_2}, \cdots, \mathbf{w_n})$ denote a sequence of tokens and $\mathbf{y}=(\mathbf{y}_0, \mathbf{y}_1, \cdots, \mathbf{y}_m)$ denote the fixed number of semantic categories. The goal is to learn a probabilistic function that takes $\mathbf{x}$ as input and outputs the probability distribution over $\mathbf{y}$. Without loss of generality, we only study the binary text classification problem under noisy setting in this work.

\subsection{A Practical Scenario}
In the context of supervised machine learning, we assume that in most scenarios, we only have access to labelled noise-free texts. There are two reasons. Firstly, most open-sourced labelled data do not consider OCR noises. Secondly, manual labelling usually also labels clean texts, and does not consider OCR noise. One reason is that annotating noisy texts is difficult or ambiguous. Another reason is that labelling becomes subject to changes in OCR recognition. For different OCR, we need to repeat the labelling multiple times.

In order to boost the performance of model when applied on OCR transcripts, we can train or finetune the model on labelled noisy data. Then the question becomes how to transform labelled noise-free texts into labelled noisy texts. Due to the fact that labelled texts do not come with corresponding images, it is impossible to call OCR engines and obtain natural OCR transcripts. Human resources can be employed to copy texts and take pictures, but it is extremely expensive considering the size of data for model training. Then the core question is how to inject natural OCR noise into labelled texts efficiently.

\subsection{OCR Noise Simulation}
When OCR engine transforms images into texts, we can think of it as a noise induction process. Let $\mathbf{I}$ denote a handwritten image, $\mathbf{x}$ denotes the text content on image $\mathbf{I}$, OCR would transform the noise-free text $\mathbf{x}$ into its noisy copy $\mathbf{x}^\prime$. 



The problem is then defined as modeling a noise induction function 
$\mathcal{\widetilde{X}}=\mathcal{F}(\mathcal{X}, \theta)$ 
where $\theta$ is the function parameters and $\mathcal{X}$ is a collection of noise-free texts. A good simulation function makes sure that the simulated $\mathcal{\widetilde{X}}$ is close to the natural OCR transcripts $\mathcal{X^\prime}$. It should be noted that noise induction should not change the semantic meaning of content so that $\mathcal{X}$, $\mathcal{X^\prime}$ and $\mathcal{\widetilde{X}}$ share the same semantic label in text classification task.

\subsection{Robust Training}
In this work, we deal with off-line handwritten text recognition. We do not study how to improve the accuracy of recognition, but only use the recognition model as a black box tool. Instead, we are interested in how to make downstream NLP models intrinsically robust to noisy inputs.

Let $\mathcal{M}$ denote a pre-trained model that is finetuned on a noise-free dataset $\mathcal{(X, Y)}$, firstly we investigate how much performance degrades when $\mathcal{M}$ is applied on natural OCR transcripts $\mathcal{X^\prime}$. Secondly, we study on how to finetune $\mathcal{M}$ on simulated noisy datasets $\mathcal{(\widetilde{X}, Y)}$ efficiently to improve its performance on input $\mathcal{X^\prime}$ that contains natural OCR errors.

\section{Approach}
\label{approach}
\subsection{OCR Noise Simulation}
\label{OCR Noise Simulation}
In this section, we introduce the multi-source noise simulation method. 

\subsubsection{Rule-based Simulation}
\label{Rule-based Simulation}
One type of frequent noise introduced by OCR engines is the token level edit. For example, a word that is not clearly written could be mistakenly recognized as other synonymous word, or in even worse case, not recognized at all. In order to synthesize token level natural OCR noise from noise-free texts, we compare and align parallel data of clean and natural OCR transcript pairs $\mathcal{(X, X^\prime)}$ using the Levenshtein distance metric (Levenshtein, 1966). Let $\mathcal{V}$ be the vocabulary of tokens, we then construct a token level confusion matrix $\mathcal{C}_{conf}$ by aligning parallel data and estimating the probability $P(\mathbf{w}^\prime|\mathbf{w})$ with the frequency of replacing token $\mathbf{w}$ to $\mathbf{w}^\prime$, where $\mathbf{w}$ and $\mathbf{w^\prime}$ are both tokens in $\mathcal{V}$. We introduce an additional token $\mathbf{\epsilon}$ into the vocabulary to model the insertion and deletion operations, the probability of insertion and deletion can then be formulated as $P_{ins}(\mathbf{w}|\mathbf{\epsilon}$) and $P_{del}(\mathbf{\epsilon}|\mathbf{w}$) respectively. For every clean sentence $\mathbf{x}=(\mathbf{w_0}, \mathbf{w_1}, \mathbf{w_2}, \cdots, \mathbf{w_n})$, we independently perturb each token in $\mathbf{x}$ with the following procedure, which is proposed by \cite{namysl2020nat}:

\begin{itemize}
\item{Insert the $\mathbf{\epsilon}$ token before the first and after every token in sentence $\mathbf{x}$ and acquire an extended version $\mathbf{x}_{ext}=(\mathbf{\epsilon}, \mathbf{w_0}, \mathbf{\epsilon}, \mathbf{w_1}, \mathbf{\epsilon}, \mathbf{w_2}, \mathbf{\epsilon}, \cdots, \mathbf{\epsilon}, \mathbf{w_n}, \mathbf{\epsilon})$.
}

\item{For every token $\mathbf{w}$ in sentence $\mathbf{x}_{ext}$, sample another token from the probability distribution $P(\mathbf{w}^\prime|\mathbf{w})$ to replace $\mathbf{w}$.}

\item{Remove all $\mathbf{\epsilon}$ tokens from the sentence to obtain the rule-based simulated noisy sentence $\mathbf{\widetilde{x}}$.}
\end{itemize}

\subsubsection{Attack-based Simulation}
The attack-based method greedily searches for the weak spots of the input sentence \cite{yang2020greedy} by replacing each word, one at a time, with a “padding” (a zero-valued vector) and examining the changes of output probability. After finding the weak spots, attack-based method replaces the original token with another token. One drawback of greedy attack is that adversarial examples are usually unnatural \cite{hsieh2019robustness}. In even worse case, the semantic meaning of the original text might change, this makes the simulated text a bad adversarial example. To avoid such problem, we only replace the original token with its synonym. The synonym comes from the confusion matrix $\mathcal{C}_{conf}$ by aligning clean texts and OCR transcripts. This effectively constrains the semantic drifts and makes the simulated texts close to natural OCR transcripts.

\subsubsection{Model-based Simulation}
We observe that there are both token level and span level noises in natural OCR transcripts. In span level noises, there are dependencies between the recognition of multiple tokens. For example, a noise-free sentence
\begin{CJK}{UTF8}{gbsn}
“乌龟默默想着”
\end{CJK}
(translated as "The tortoise meditated" by Google Translate \footnote{https://translate.google.cn}) is recognised as
\begin{CJK}{UTF8}{gbsn}
"乌乌黑黑的箱子"
\end{CJK}
(translated as "Jet black box" by Google Translate). 
A possible reason is that the mis-recognition of 
\begin{CJK}{UTF8}{gbsn}
“龟”
\end{CJK}
leads to recognizing  
\begin{CJK}{UTF8}{gbsn}
“默”
\end{CJK}
into 
\begin{CJK}{UTF8}{gbsn}
“黑”
\end{CJK}
because 
\begin{CJK}{UTF8}{gbsn}
“乌黑”
\end{CJK}
is a whole word in Chinese.
The rule-based and attack-based simulation mainly focuses on token-level noise where a character or token might be edited. It makes edits independently and does not consider dependency between multiple tokens. As a consequence, both ruled-based and attack-based simulation are not able to synthesize the span level noise. 

We proposed to model both token level and span level noise using the encoder-decoder architecture, which is successful in many NLP tasks such as machine translation, grammatical error corrections (GEC) and etc.. While a GEC model takes noisy texts as input and generates noise-free sentences, our model-based noise injection model is quite the opposite. During training, we feed parallel data of clean and OCR transcripts ($\mathcal{X}$, $\mathcal{X^\prime}$) into the injection model so that it can learn the frequent errors that OCR engines will make. During inference, the encoder first encode noise-free text into a fix length representation and the decoder generates token one step a time with possible noise in an auto-regressive manner. This makes sure that both token level and span level noise distribution can be captured by the model. We can use the injection model to synthesize a large number of noisy texts that approximate the natural OCR errors. It should be noted that the injection model is not limited to a certain type of encoder-decoder architecture. In our experiment, we employ a 6-layer vanilla Transformer (base model) as in \cite{vaswani2017attention}.

\subsection{Noise Invariance Representation}
\cite{zheng2016improving} pointed out the output instability issues of deep neural networks. They presented a general stability training method to stabilize deep networks against small input distortions that result from various types of common image processing. Inspired by \cite{zheng2016improving}, \cite{namysl2020nat} adapted the stability training method to the sequence labeling scenario. Here we adapt it to the text classification task.
Given the standard task objective $\mathcal{L}_{stand}$, the clean text $\mathbf{x}$, its simulated noisy copy $\mathbf{\widetilde{x}}$ and the shared label $\mathbf{y}$, the stability loss is defined as

\begin{equation}
\label{eq:total_loss}
\mathcal{L}=\alpha *\mathcal{L}_{stand}+(1-\alpha) * \mathcal{L}_{sim}
\nonumber
\end{equation}

\begin{equation}
\label{eq:sim_loss}
\mathcal{L}_{sim}=Distance(\mathbf{y(x}), \mathbf{y(\widetilde{x}}))
\nonumber
\end{equation}

where $\mathcal{L}_{sim}$ is the distance between model outputs for clean input $\mathbf{x}$ and noisy input $\mathbf{\widetilde{x}}$, $\alpha$ is a hyper-parameter to trade off $\mathcal{L}_{stand}$ and $\mathcal{L}_{sim}$. $\mathcal{L}_{sim}$ is expected to be small so that the model is not sensitive to the noise disturbance. This enables the model to obtain robust representation for both clean and noisy input. Specifically, we use cosine distance as our distance measure.

\subsection{Hard Example Mining}
The proposed noise simulation methods could generate quadratic or cubic number of parallel samples compared to the size of original dataset. It is good that we now have sufficient number of training data with noises and labels. Nevertheless, the training process becomes inefficient if we naively treat each simulated sample equally and feed all the samples into the classifier. This makes the training process extremely time-consuming and does not lead to an optimal performance. Consequently, we need a strategy to sample examples from large volumes of data for optimal performance.
Ideally, a robust model should learn similar representations for all possible noise-free text $\mathbf{x}$ and its corresponding noisy copy $\mathbf{\widetilde{x}}$. In reality, however, the model can only capture noise-invariance representations for some of the simulated samples, for some other samples, the representations of the clean text and its noisy copy are still quite different. 
For any given model $\mathcal{M}$, we define a sample $\mathbf{x}$ as a hard example for $\mathcal{M}$ if the representations of $\mathbf{x}$ and $\mathbf{\widetilde{x}}$ are not similar. We believe that at different training iterations, the hard examples are different, and the model should focus more on the hard ones.
We propose a hard example mining algorithm that dynamically distinguishes hard and easy samples for each training epoch as follows:
\begin{itemize}
    \item{Step 1. Initialize the classifier by finetuning a pre-trained model on the noise-free training data $\mathcal{D}_{clean}=\{\mathbf{x}_i\}_{i=1,2,...N}$}
    \item{Step 2. Generate a large number of simulated noisy texts $\mathcal{D}_{noisy}=\{\mathbf{\widetilde{x}}_i\}_{i=1,2,...M}$ and construct a collection of all training samples $\mathcal{D}=\{\mathcal{D}_{clean}, \mathcal{D}_{noisy}\}$}
    \item{Step 3. For each iteration $t$, we feed training samples $\mathcal{D}$ to the classifier and obtain their representations $\mathcal{E}_t=\{\mathbf{e}_i, \mathbf{\widetilde{e}}_i\}_{i=1,2,...M}$ from classifier.}
    \item{Step 4. Calculate the cosine distance of $\mathbf{e}_i$ and $\mathbf{\widetilde{e}}_i$. Rank all the distances, i.e., $Distance = \{cosine(\mathbf{e}_i, \mathbf{\widetilde{e}}_i)\}_{i=1,2,...M}$, and only keep samples with the top largest distance. These are the hard examples and we use $\mathcal{D}_{hard}$ to denote it. We use a hyper-parameter $\beta=|{D}_{hard}|/M$ to control the number of hard examples.}
    \item{Step 5. Train classifier on $\mathcal{D}_t$=\{$\mathcal{D}_{hard}$, $\mathcal{D}_{clean}$\} and update model by miniming $\mathcal{L}=\alpha *\mathcal{L}_{stand}+(1-\alpha) * \mathcal{L}_{sim}$}
\end{itemize}

\subsection{The Overall Framework}
\begin{figure*}[!tpbh]
\centering
\includegraphics[width=0.8\textwidth] {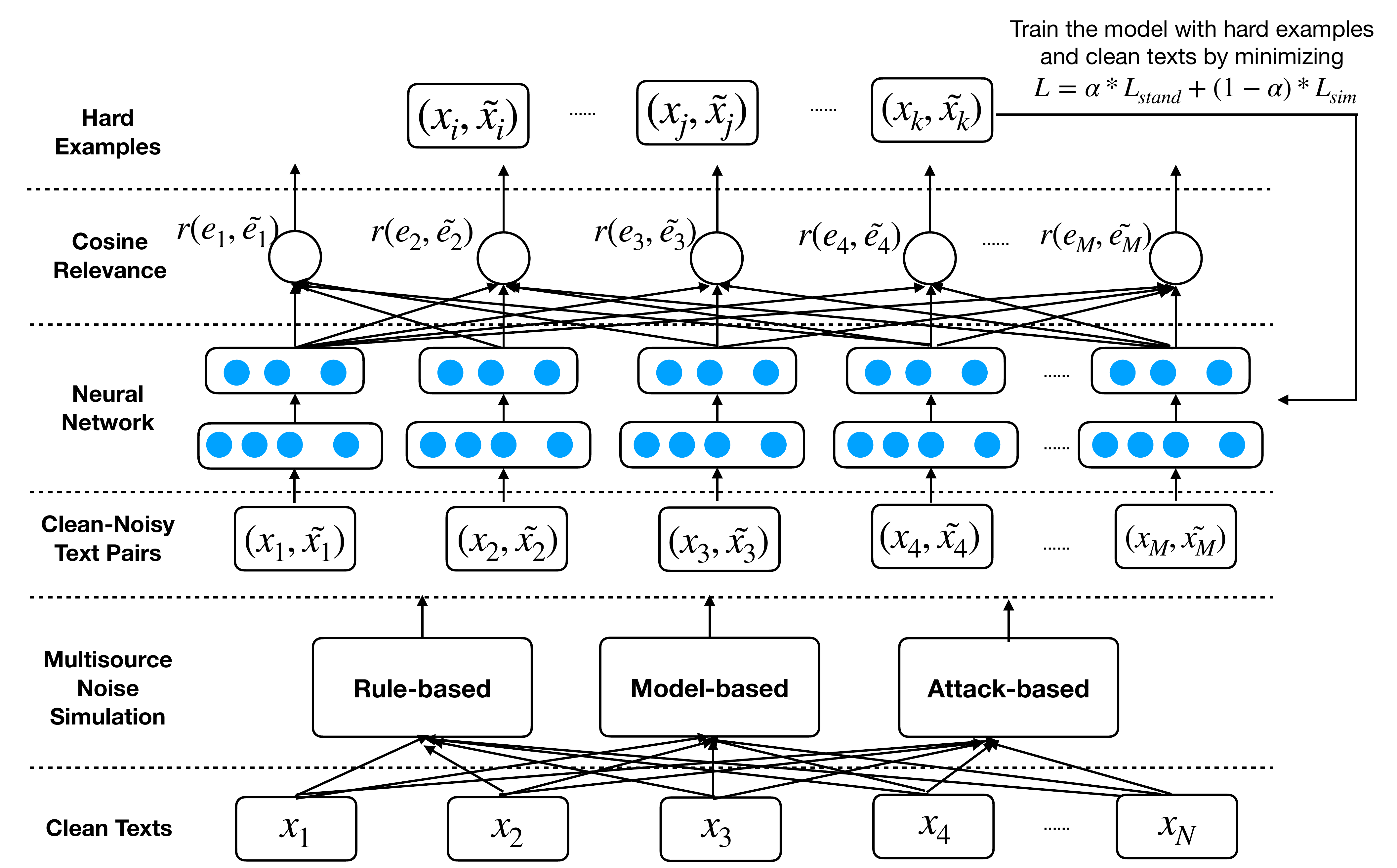}
\caption{The overview of the robust training framework.}
\label{fig:framework}
\end{figure*}

The overall framework is shown in Figure \ref{fig:framework}. Let $\mathbf{x}_i$, $i=0, 1, 2,...N$ denote the noise-free text, where $\mathbf{x}_i$ is a sequence of tokens, and $\mathbf{\widetilde{x}}_i$ is the simulated noisy copy. $\mathbf{e}_i$ and $\mathbf{\widetilde{e}}_i$ are the model representations for $\mathbf{x}_i$ and $\mathbf{\widetilde{x}}_i$, we calculate the cosine distance between $\mathbf{e}_i$ and $\mathbf{\widetilde{e}}_i$ and select those pairs with largest distance as the hard examples. Then hard examples together with original noise-free data are used to train the model. For each iteration, we select hard examples dynamically.

\section{Experiment}
\label{experiment}
\subsection{Dataset}
We describe three evaluation datasets and the parallel data for training the model-based noise simulation model below.
\subsubsection{Test Data}
To comprehensively evaluate pre-trained models and the proposed framework, we perform experiments on three real-world text classification datasets, e.g., Metaphor, Personification and Parallelism detection. In each dataset, the task is to assign a positive or negative label to a sentence.
\begin{itemize}
    \item{Metaphor, is a figure of speech that describes an object or action in a way that is not literally true, but helps explain an idea or make a comparison.}
    \item{Personification, is a figure of speech when you give an animal or object qualities or abilities that only a human can have.}
    \item{Parallelism, is a figure of speech when phrases in a sentence have similar or the same grammatical structure.}
\end{itemize}
In order to obtain the above three datasets, we hired five professional teachers to annotate essays of primary school students. We broke down essays into sentences and each sentence was annotated as one of the three rhetoric or did not belong to any rhetoric. We aggregated crowd-sourced labels into ground-truth label using majority voting. Each task contains over 2000 sentences and the number of positive examples is between 48 to 156. It should be noted that this imbalance is caused by the fact that rhetoric is not so common in students' essays. We simply keep the natural positive and negative sample ratio in the test set for objectiveness.
Details about the test data are in Table \ref{tab:testdataStats}.

\subsubsection{OCR Engine and Natural Noise}
Different from existing work \cite{namysl2020nat} which evaluated model performance on simulated OCR transcripts, we constructed six real OCR test data for evaluation. We hired over 20 people to write down the original noise-free texts, take pictures and feed images to commercial OCR engines so that natural OCR transcripts can be obtained. We chose Hanvon OCR \footnote{https://www.hw99.com/index.php} and TAL OCR \footnote{https://ai.100tal.com/product/ocr-hr} as our engines because they are the leading solutions for Chinese primary school student's handwriting recognition. The noise rates are 3.42\% and 6.11\% for Hanvon and TAL OCR test data respectively. Because we can only experiment with limited number of OCR engines, we discuss the impact of different noise levels in section 6.1.

\begin{table}
 \setlength{\tabcolsep}{1.6mm}
\centering
 \caption{Test data.}
\begin{tabular}{| c | c | c | c | c |}
\hline
  Dataset & \#sentences & \#positives & AvgSentLen\\
\hline
  Metaphor         & 2064 & 156 & 37.5 \\
  Personification      & 2059 & 64 & 37.6 \\
  Parallelism         & 2063 & 48  & 37.5 \\ 
\hline
\end{tabular}
 \label{tab:testdataStats}
\end{table}

\subsubsection{Parallel Data for Noise Simulation}
In order to train the model-based noise simulation model in section 4.1.3, we collect about 40,000 parallel data\footnote{Parallel data do not have task specific labels, so they are not used as training data} of human transcripts and OCR transcripts as our training data. We believe that 40,000 is a reasonable amount to train a high quality model-based noise generator. More importantly, once trained, the model can serve as a general noise generator regardless of specific tasks. In other words, we can use it to quickly convert annotated clean text into annotated noisy text in all sorts of tasks.

\subsection{Implementation}
For each classification task, we first finetune pre-trained models on noise-free training data $\mathcal{D}_{clean}$, save models with the best validation loss as $\mathcal{M^*}_{clean}$. To perform robust training, we synthesize noisy copies of the original training data and then finetune $\mathcal{M^*}_{clean}$ on both clean and noisy data as denoted by $\mathcal{M^*}_{noisy}$. Both $\mathcal{M^*}_{clean}$ and $\mathcal{M^*}_{noisy}$ are tested on original noise-free test data and noisy copies of the test data.

We implement the framework using PyTorch and train models on Tesla V100 GPUs. 
We use an opensource release\footnote{https://github.com/ymcui/Chinese-BERT-wwm} of Chinese BERT and RoBERTa as the pre-trained models. We tune learning rate $\in \{5e^{-8}, 5e^{-7}\}$, batch size $\in \{5, 10\}$, $\alpha \in \{1.0, 0.75, 0.50 \}$ where $\alpha=1.0$ indicates no stability loss is employed. We keep all other hyper-parameters as they are in the release. We report precision, recall and F1 score as performance metrics.

\subsection{Results}
\subsubsection{Robust Training on Simulated Texts}
Instead of naively combining multi-source simulation data and finetuning model $\mathcal{M^*}_{clean}$ on it, we employ the hard example mining algorithm in section 4.3 and the stability loss in section 4.2 for robust training. We compare the proposed robust training framework with several strong baselines.
\begin{itemize}
    \item {Random. We randomly select several tokens and make insertion, deletion or substitution edits to generate permuted data. We then combine the permuted and clean data and finetune models on it.}
    \item{Noise-aware Training, i.e., NAT \cite{namysl2020nat}, noise-aware training for robust neural sequence labeling, which proposes two objectives, namely data augmentation and stability loss, to improve the model robustness in perturbed input.}
    \item{TextFooler, \cite{jin2020bert}, a strong baseline to generate adversarial text for robust adversarial training.}
    \item{Naively Merge. We finetune $\mathcal{M^*}_{clean}$ on clean and noisy samples generated by all three simulation methods, but without hard example mining and stability loss.}
\end{itemize}
The results are in Table \ref{tab:main}. Ours is the proposed robust training framework that finetunes $\mathcal{M^*}_{clean}$ on clean and noisy samples generated by all three simulation methods, together with hard example mining and stability loss. We have the following observations:
\begin{itemize}
\item{Compared with $\mathcal{M}^*_{clean}$, all robust training approaches, Random, NAT, TextFooler, Naively Merge and our robust training framework (Ours) improve the F1 score on both noise-free test data and OCR test data on all three tasks.}
\item{Compared with Naively Merge, Ours demonstrates improvements in both precision and recall in all test data, which proves that hard example mining and stability loss are vital to the robust training framework.}
\item{When compared with existing baselines, Ours ranks the first place eight times and the second place once out of all nine F1 scores (three tasks, three test data for each task). This proves the advantages of using the proposed robust training framework over existing approaches.}
\end{itemize}
  We think of two reasons. Firstly, the proposed noise simulation method generates more natural noisy samples than baselines do. Baselines might introduce plenty of unnatural noisy samples, making precision even lower that of $\mathcal{M^*}_{clean}$. Secondly, hard example mining algorithm enables the model to focus on hard examples whose robust representation has not been learned. NAT and TextFooler finetunes models by naively combing clean and noisy samples.

\begin{table}[!ht]
\centering
 \caption{Evaluation results of BERT on metaphor, personification and parallelism.}
   \begin{threeparttable}       
   
 \setlength{\tabcolsep}{1.3mm}
\begin{tabular}{r*{10}{c}}
  \toprule
  && \multicolumn{3}{c}{Noise-free Data} & \multicolumn{3}{c}{Hanvon OCR} & \multicolumn{3}{c}{TAL OCR}\\
   \cmidrule(lr){3-5} \cmidrule(lr){6-8} \cmidrule(lr){9-11}
   & Task & P & R & F1 & P & R & F1 & P & R & F1\\
  \midrule
  $\mathcal{M^*}_{clean}$   & Metaphor & \bf{0.897} & 0.833 & 0.864 & 0.888 & 0.814 & 0.849 & \bf{0.886} & 0.795 & 0.838 \\
  Random                    & Metaphor & 0.873 & \bf{0.885} & 0.879 & 0.864 & 0.853 & 0.858 & 0.868 & 0.840 & 0.854 \\
  NAT                       & Metaphor & 0.871 & 0.866 & 0.868 & 0.868 & 0.846 & 0.857 & 0.877 & 0.821 & 0.848 \\
  TextFooler                & Metaphor & 0.883  & 0.872 & 0.877 & 0.874 & 0.846 & 0.860 & 0.872 & 0.833 & 0.852 \\ 
  Naively Merge             & Metaphor & 0.877 & 0.872 & 0.875 & 0.880 & 0.846 & 0.863 & 0.873 & 0.833 & 0.852 \\
  Ours        	            & Metaphor & 0.890 & \bf{0.885} & \bf{0.887} & \bf{0.889} & \bf{0.872} & \bf{0.880} & 0.877 & \bf{0.865} & \bf{0.871} \\ \midrule
  $\mathcal{M^*}_{clean}$   & Personification & 0.855 & 0.828 & 0.841 & 0.868 & 0.719 & 0.787 & 0.825 & 0.734 & 0.777 \\
  Random                    & Personification & 0.831 & \bf{0.844} & 0.837 & 0.814 & 0.750 & 0.781 & 0.842 & 0.750 & 0.793 \\
  NAT                       & Personification & 0.925 & 0.766 & 0.838 & 0.904 & 0.734 & 0.810 & 0.917 & 0.688 & 0.786 \\
  TextFooler                & Personification & 0.831  & \bf{0.844} & 0.837 & 0.803 & \bf{0.766} & 0.784 & 0.831 & \bf{0.766} & 0.797 \\ 
 Naively Merge              & Personification & 0.895 & 0.797 & 0.843 & 0.875 & 0.766 & 0.817 & 0.885 & 0.719 & 0.793 \\
  Ours        	            & Personification & \bf{0.927} & 0.797 & \bf{0.857} & \bf{0.923} & 0.750 & \bf{0.828} & \bf{0.926} & 0.734 & \bf{0.817} \\ \midrule
  $\mathcal{M^*}_{clean}$   & Parallelism & 0.720 & 0.750 & 0.735 & 0.756 & 0.646 & 0.697 & 0.725 & 0.604 & 0.659 \\
  Random                    & Parallelism & 0.717 & \bf{0.792} & 0.753 & 0.714 & \bf{0.729} & 0.721 & 0.717 & 0.688 & 0.702 \\
  NAT                       & Parallelism & \bf{0.814} & 0.729 & \bf{0.769} & \bf{0.821} & 0.667 & 0.736 & \bf{0.795} & 0.646 & 0.713 \\
  TextFooler                & Parallelism & 0.731  & \bf{0.792} & 0.760 & 0.733 & 0.688 & 0.710 & 0.767 & 0.688 & 0.725 \\ 
Naively Merge               & Parallelism & 0.777 & 0.729 & 0.753 & 0.781 & 0.667 & 0.719 & 0.781 & 0.667 & 0.719 \\
  Ours        	            & Parallelism & 0.783 & 0.750 & 0.766 & 0.773 & 0.708 & \bf{0.739} & 0.778 & \bf{0.729} & \bf{0.753} \\
  \bottomrule
\end{tabular}
 \label{tab:main}
 
  \end{threeparttable}       
  
\end{table}

\section{Analysis}
\label{analysis}
\subsection{Naive Training with A Single Noise Simulation Method}
We introduce our multi-source noise simulation methods in section 4.1. Using these methods, we can generate a large number of noisy texts from noise-free data. In this section, we evaluate the effectiveness for each method independently. We reload $\mathcal{M^*}_{clean}$ and finetune it combining clean texts and noisy texts generated by a single noise simulation method. At this stage, neither hard example mining nor stability loss is employed. The results of using a single noise simulation method are listed in Tables \ref{tab:sim@metaphor}, \ref{tab:sim@personification}, \ref{tab:sim@parallel}. $\mathcal{M^*}_{clean}$ is finetuned on noise-free data. Rule-based, Model-based and Attack-based are finetuned with a single noise simulation method without hard example mining and stability loss


Firstly, we observe that both recall and F1 score decrease significantly on two noisy test sets compared to performance on noise-free test set. For example, on TAL OCR test set, F1 score of BERT decreases 6.4\% and 7.6\% for Personification and Parallelism detection and F1 score of RoBERTa decreases 8.5\% and 4.0\% respectively. This proves that pre-trained models trained on noise-free data are not robust to OCR noises.

Secondly, all three noise simulation methods can improve the F1 scores of BERT and RoBERTa for all three tasks. However, when we naively combine multi-source simulations and finetune models on it (``Naively Merge" in Table 2), the performance does not exceed the effect of using a single noise simulation method. This motivates us to introduce hard example mining and stability loss into the proposed robust training framework.

\begin{table}[!ht]
\centering
 \caption{Performance on metaphor detection with a single noise simulation.}
 \setlength{\tabcolsep}{1.5mm}
 \begin{threeparttable}          
\begin{tabular}{r*{10}{c}}
  \toprule
  && \multicolumn{3}{c}{Noise-free Data} & \multicolumn{3}{c}{Hanvon OCR} & \multicolumn{3}{c}{TAL OCR}\\
  \cmidrule(lr){3-5} \cmidrule(lr){6-8} \cmidrule(lr){9-11}
  Simulation & Model & P & R & F1 & P & R & F1 & P & R & F1\\
  \midrule
 $\mathcal{M^*}_{clean}$ & BERT & \bf{0.897} & 0.833 & 0.864 & \bf{0.888} & 0.814 & 0.849 & 0.886 & 0.795 & 0.838 \\
  Rule-based            & BERT & 0.877 & \bf{0.872} & \bf{0.874} & 0.874 & \bf{0.846} & 0.860 & 0.872 & \bf{0.833} & \bf{0.852} \\
  Model-based           & BERT & 0.882  & 0.865 & 0.873 & 0.885 & 0.840 & \bf{0.862} & 0.878 & 0.827 & \bf{0.852} \\ 
  Attack-based        	& BERT & 0.887 & 0.859 & 0.873 & 0.879 & 0.840 & 0.859 & \bf{0.894} & 0.808 & 0.849 \\ \midrule
 $\mathcal{M^*}_{clean}$ & RoBERTa & 0.872 & \bf{0.917} & \bf{0.894} & 0.862 & 0.878 & 0.870 & \bf{0.873} & 0.878 & 0.875 \\
  Rule-based            & RoBERTa & 0.836 & \bf{0.917} & 0.875 & 0.821 & 0.910 & 0.863 & 0.844 & \bf{0.904} & 0.873 \\
  Model-based           & RoBERTa & 0.872 & \bf{0.917} & \bf{0.894} & 0.856 & \bf{0.917} & \bf{0.885} & 0.859 & 0.897 & \bf{0.878} \\ 
  Attack-based        	& RoBERTa & \bf{0.889} & 0.872 & 0.880 & \bf{0.879} & 0.840 & 0.859 & 0.872 & 0.827 & 0.849 \\
 \bottomrule
\end{tabular}

 \label{tab:sim@metaphor}
 \end{threeparttable}       
\end{table}

\begin{table}[!ht]
\centering
 \caption{Performance on personification detection with a single noise simulation.}
  \begin{threeparttable}          
 \setlength{\tabcolsep}{1.5mm}
\begin{tabular}{r*{10}{c}}
  \toprule
  && \multicolumn{3}{c}{Noise-free Data} & \multicolumn{3}{c}{Hanvon OCR} & \multicolumn{3}{c}{TAL OCR}\\
  \cmidrule(lr){3-5} \cmidrule(lr){6-8} \cmidrule(lr){9-11}
  Simulation & Model & P & R & F1 & P & R & F1 & P & R & F1\\
  \midrule
 $\mathcal{M^*}_{clean}$         & BERT & 0.855 & 0.828 & 0.841 & \bf{0.868} & 0.719 & 0.787 & 0.825 & 0.734 & 0.777 \\
  Rule-based            & BERT & 0.818 & \bf{0.844} & 0.831 & 0.817 & \bf{0.766} & 0.791 & 0.833 & \bf{0.781} & \bf{0.806} \\
  Model-based           & BERT & \bf{0.862}  & 0.781 & 0.820 & 0.855 & 0.734 & 0.790 & \bf{0.855} & 0.734 & 0.790 \\ 
  Attack-based        	& BERT & 0.844 & \bf{0.844} & \bf{0.844} & 0.831 & \bf{0.766} & \bf{0.797} & 0.831 & 0.765 & 0.797 \\ \midrule
 $\mathcal{M^*}_{clean}$          & RoBERTa & 0.764 & \bf{0.859} & 0.809 & 0.754 & 0.812 & 0.782 & 0.730 & 0.719 & 0.724 \\
  Rule-based            & RoBERTa & 0.775 & \bf{0.859} & 0.815 & 0.783 & \bf{0.844} & \bf{0.812} & 0.739 & \bf{0.797} & 0.767 \\
  Model-based           & RoBERTa & 0.776 & 0.812 & 0.794 & 0.785 & 0.797 & 0.791 & \bf{0.817} & 0.766 & \bf{0.791} \\ 
  Attack-based        	& RoBERTa & \bf{0.850} & 0.797 & \bf{0.823} & \bf{0.828} & 0.750 & 0.787 & 0.808 & 0.656 & 0.724 \\
  \bottomrule
\end{tabular}


 \label{tab:sim@personification}
  \end{threeparttable}       
\end{table}

\begin{table}[!ht]
\centering
 \caption{Performance on parallelism detection with a single noise simulation.}
  \begin{threeparttable}          
 \setlength{\tabcolsep}{1.5mm}
\begin{tabular}{r*{10}{c}}
  \toprule
  && \multicolumn{3}{c}{Noise-free Data} & \multicolumn{3}{c}{Hanvon OCR} & \multicolumn{3}{c}{TAL OCR}\\
  \cmidrule(lr){3-5} \cmidrule(lr){6-8} \cmidrule(lr){9-11}
  Simulation & Model & P & R & F1 & P & R & F1 & P & R & F1\\
  \midrule
 $\mathcal{M^*}_{clean}$         & BERT & 0.720 & 0.750 & 0.735 & 0.756 & 0.646 & 0.697 & 0.725 & 0.604 & 0.659 \\
  Rule-based            & BERT & 0.679 & \bf{0.792} & 0.731 & 0.700 & \bf{0.714} & \bf{0.758} & 0.739 & \bf{0.708} & 0.723 \\
  Model-based           & BERT & \bf{0.771}  & 0.771 & \bf{0.771} & 0.733 & 0.688 & 0.710 & 0.786 & 0.688 & \bf{0.734} \\ 
  Attack-based        	& BERT & 0.766 & 0.750 & 0.758 & \bf{0.789} & 0.625 & 0.698 & \bf{0.800} & 0.667 & 0.727 \\ \midrule
 $\mathcal{M^*}_{clean}$ & RoBERTa & \bf{0.795} & 0.729 & 0.761 & \bf{0.838} & 0.646 & 0.730 & \bf{0.816} & 0.646 & 0.721 \\
  Rule-based            & RoBERTa & 0.780 & \bf{0.812} & \bf{0.796} & 0.800 & \bf{0.750} & \bf{0.774} & 0.795 & \bf{0.729} & \bf{0.761} \\
  Model-based           & RoBERTa & 0.792 & 0.792 & 0.792 & 0.814 & 0.729 & 0.769 & 0.810 & 0.708 & 0.756 \\ 
  Attack-based        	& RoBERTa & 0.787 & 0.771 & 0.779 & 0.829 & 0.708 & 0.764 & 0.805 & 0.688 & 0.742 \\
  \bottomrule
  \end{tabular}
 \label{tab:sim@parallel}
 
 
  \end{threeparttable}       
\end{table}

\subsection{The Impact of Different Noise Level}
\begin{figure}[!ht]
\centering
\includegraphics[width=0.8\textwidth] {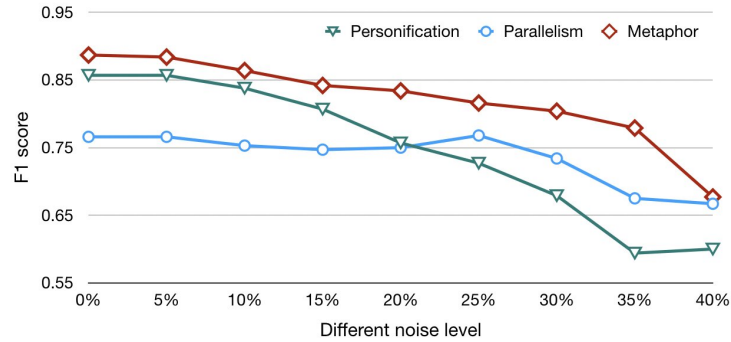}
\caption{The impact of different noise levels.}
\label{fig:different_noise_level}
\end{figure}

We prove that the proposed robust training framework can largely boost model performance when applied on noisy inputs generated by real OCR engines. Since the noise rate in both Hanvon and TAL OCR test data is relatively low, we have not evaluated the effectiveness of the proposed robust training framework under different noise rates, especially when there are significant number of noises in the inputs. In this section, we investigate this problem and show the results in Figure \ref{fig:different_noise_level}. We introduce different levels of noises by randomly inserting, deleting or replacing tokens in noise-free texts with equal probability.

As shown in Figure \ref{fig:different_noise_level}, we can observe that F1 score decreases as the noise rate increases. When noise rate is less than 25\%, F1 score decreases slowly for Parallelism and Metaphor detection, and drops significantly when noise rate exceeds 30\%. Another observation is that performance of Personification detection degrades faster than the other two tasks, as reflected in a sharper slope in Figure \ref{fig:different_noise_level}.

\begin{figure*}[!ht]
\centering
\includegraphics[width=0.9\textwidth]{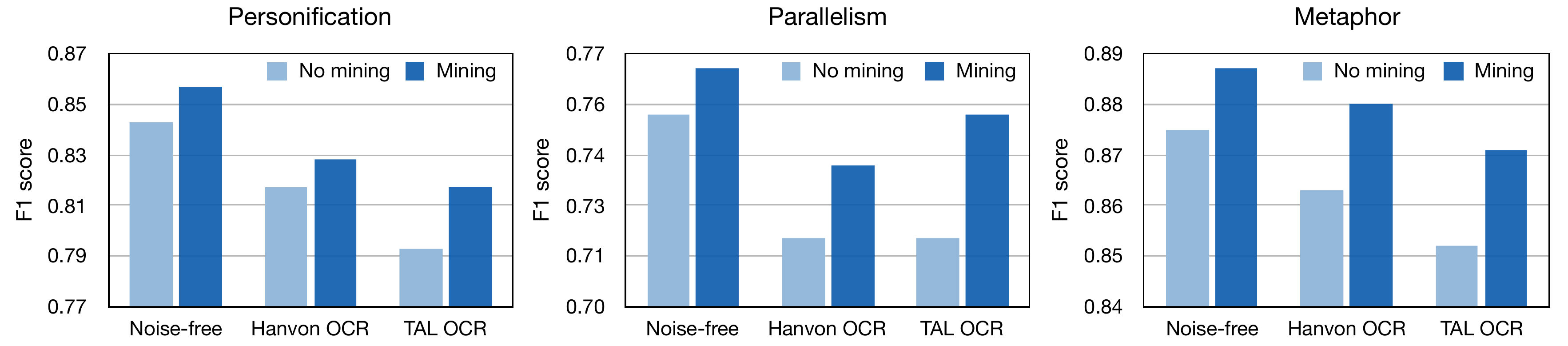}
\caption{The impact of hard example mining.}
\label{fig:impact_mining}
\end{figure*}

\begin{figure*}[!ht]
\centering
\includegraphics[width=0.9\textwidth] {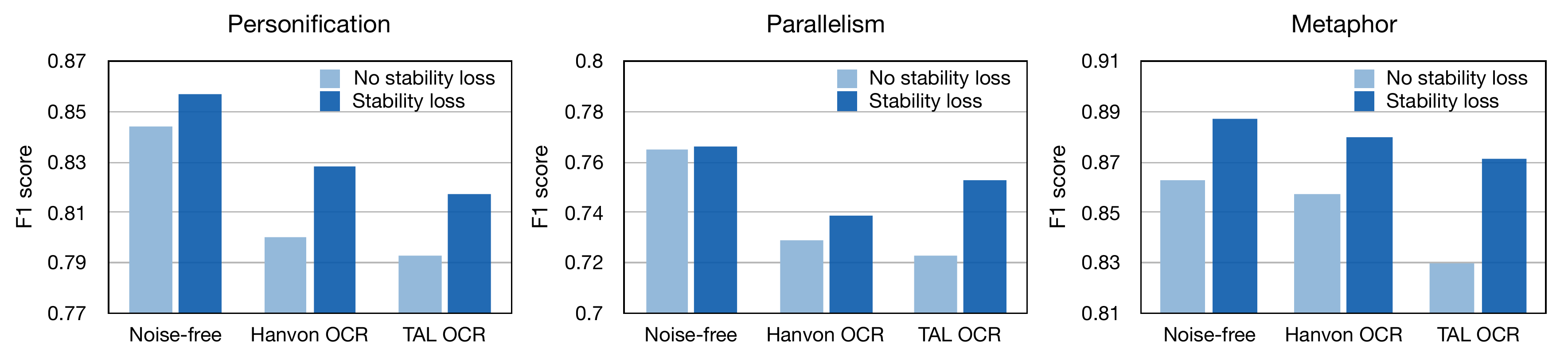}
\caption{The impact of stability loss.}
\label{fig:impact_stability}
\end{figure*}

\subsection{The Impact of Hard Example Mining}
Hard example mining algorithm allows the model to dynamically pay more attention to hard examples $(\mathbf{x_i}, \mathbf{\widetilde{x_i}})$ whose representations $(\mathbf{e_i}, \mathbf{\widetilde{e_i}})$ are still quite different. We believe that it is vital for the model to learn robust representations. In this section, we investigate the performance difference with and without hard example mining. As shown in Figure \ref{fig:impact_mining}, F1 score consistently increases for both noise-clean and noisy OCR test data when hard example mining is employed. For example, hard example mining improves F1 by 2\% on Metaphor and 3.4\% on Parallelism using TAL OCR. This indicates the importance of hard example mining in the proposed framework.

\subsection{The Impact of Stability Loss}
The use of stability loss guarantees that model can learn similar representations for clean text $\mathbf{x}$ and its noisy copy $\mathbf{x^\prime}$. In this section, we investigate the performance difference with and without stability loss. As shown in Figure \ref{fig:impact_stability}, F1 score decreases when there are no stability loss for all three datasets. On Metaphor detection, using stability loss improves F1 by 4.1\% and 2.3\% for TAL OCR and Hanvon OCR. This indicates that stability loss is vital to the proposed framework.

\section{Conclusion}
\label{conclusion}

In this paper, we study the robustness of multiple pre-trained models, e.g., BERT and RoBERTa, in text classification when inputs contain natural OCR noises. We propose a multi-source noise simulation method that can generate both token-level and span-level noises. We finetune models on both clean and simulated noisy data and propose a hard example mining algorithm so that during each training iteration, the model can focus on hard examples whose robust representations have not been learned. For evaluation, we construct three real-world text classification datasets and obtain natural OCR transcripts by calling OCR engines on real handwritten images. Experiments on three datasets proved that the proposed robust training framework largely boosts the model performance for both clean texts and natural OCR transcripts. It also outperforms all existing robust training approaches. In order to fully investigate the effectiveness of the framework, we evaluate it under different levels of noises and study the impact of hard example mining and stability loss independently. In the future, we will experiment the proposed framework on other NLP tasks and more languages. In the meanwhile, we will study the problem under automatic speech recognition (ASR) transcripts.

\section*{Acknowledgment}
\label{sec:acknowledgement}
This work was supported in part by National Key R\&D Program of China, under Grant No. 2020AAA0104500 and in part by Beijing Nova Program (Z201100006820068) from Beijing Municipal Science \& Technology Commission.

\bibliographystyle{splncs04}
\bibliography{ecml2021}

\begin{thebibliography}{10}
\providecommand{\url}[1]{\texttt{#1}}
\providecommand{\urlprefix}{URL }
\providecommand{\doi}[1]{https://doi.org/#1}

\bibitem{alzantot-etal-2018-generating}
Alzantot, M., Sharma, Y., Elgohary, A., Ho, B.J., Srivastava, M., Chang, K.W.:
  Generating natural language adversarial examples. In: Proc. of EMNLP. pp.
  2890--2896 (2018)

\bibitem{belinkov2017synthetic}
Belinkov, Y., Bisk, Y.: Synthetic and natural noise both break neural machine
  translation. In: Proc. of ICLR (2018)

\bibitem{chollampatt2018neural}
Chollampatt, S., Ng, H.T.: Neural quality estimation of grammatical error
  correction. In: Proc. of EMNLP. pp. 2528--2539 (2018)

\bibitem{Devlin2019BERTPO}
Devlin, J., Chang, M.W., Lee, K., Toutanova, K.: {BERT}: Pre-training of deep
  bidirectional transformers for language understanding. In: Proc. of
  NAACL-HLT. pp. 4171--4186 (2019)

\bibitem{Ebrahimi2018HotFlipWA}
Ebrahimi, J., Rao, A., Lowd, D., Dou, D.: {H}ot{F}lip: White-box adversarial
  examples for text classification. In: Proc. of ACL. pp. 31--36 (2018)

\bibitem{goodfellow2014explaining}
Goodfellow, I.J., Shlens, J., Szegedy, C.: Explaining and harnessing
  adversarial examples. In: Proc. of ICLR (2015)

\bibitem{hsieh2019robustness}
Hsieh, Y.L., Cheng, M., Juan, D.C., Wei, W., Hsu, W.L., Hsieh, C.J.: On the
  robustness of self-attentive models. In: Proc. of ACL. pp. 1520--1529 (2019)

\bibitem{jin2020bert}
Jin, D., Jin, Z., Zhou, J.T., Szolovits, P.: Is {BERT} really robust? {A}
  strong baseline for natural language attack on text classification and
  entailment. In: The Thirty-Fourth {AAAI} Conference on Artificial
  Intelligence, {AAAI} 2020, The Thirty-Second Innovative Applications of
  Artificial Intelligence Conference, {IAAI} 2020, The Tenth {AAAI} Symposium
  on Educational Advances in Artificial Intelligence, {EAAI} 2020, New York,
  NY, USA, February 7-12, 2020. pp. 8018--8025 (2020)

\bibitem{karpukhin2019training}
Karpukhin, V., Levy, O., Eisenstein, J., Ghazvininejad, M.: Training on
  synthetic noise improves robustness to natural noise in machine translation.
  In: Proceedings of the 5th Workshop on Noisy User-generated Text (W-NUT
  2019). pp. 42--47 (2019)

\bibitem{miyato2016adversarial}
Miyato, T., Dai, A.M., Goodfellow, I.J.: Adversarial training methods for
  semi-supervised text classification. In: Proc. of ICLR (2017)

\bibitem{namysl2020nat}
Namysl, M., Behnke, S., K{\"o}hler, J.: {NAT}: Noise-aware training for robust
  neural sequence labeling. In: Proc. of ACL. pp. 1501--1517 (2020)

\bibitem{ndiaye2003spell}
Ndiaye, M., Faltin, A.V.: A spell checker tailored to language learners.
  Computer Assisted Language Learning (2-3),  213--232 (2003)

\bibitem{rawlinson2007significance}
Rawlinson, G.: The significance of letter position in word recognition. IEEE
  Aerospace and Electronic Systems Magazine (1),  26--27 (2007)

\bibitem{ribeiro2018semantically}
Ribeiro, M.T., Singh, S., Guestrin, C.: Semantically equivalent adversarial
  rules for debugging {NLP} models. In: Proc. of ACL. pp. 856--865 (2018)

\bibitem{sun2020adv}
Sun, L., Hashimoto, K., Yin, W., Asai, A., Li, J., Yu, P., Xiong, C.: Adv-bert:
  Bert is not robust on misspellings! generating nature adversarial samples on
  bert. arXiv preprint arXiv:2003.04985  (2020)

\bibitem{sun2019contextual}
Sun, Y., Jiang, H.: Contextual text denoising with masked language models.
  arXiv preprint arXiv:1910.14080  (2019)

\bibitem{sang2003introduction}
Tjong Kim~Sang, E.F.: Introduction to the {C}o{NLL}-2002 shared task:
  Language-independent named entity recognition. In: {COLING}-02: The 6th
  Conference on Natural Language Learning 2002 ({C}o{NLL}-2002) (2002)

\bibitem{valenti2003overview}
Valenti, S., Neri, F., Cucchiarelli, A.: An overview of current research on
  automated essay grading. Journal of Information Technology Education:
  Research (1),  319--330 (2003)

\bibitem{vaswani2017attention}
Vaswani, A., Shazeer, N., Parmar, N., Uszkoreit, J., Jones, L., Gomez, A.N.,
  Kaiser, L., Polosukhin, I.: Attention is all you need. In: Advances in Neural
  Information Processing Systems 30: Annual Conference on Neural Information
  Processing Systems 2017, December 4-9, 2017, Long Beach, CA, {USA}. pp.
  5998--6008 (2017)

\bibitem{yang2020greedy}
Yang, P., Chen, J., Hsieh, C.J., Wang, J.L., Jordan, M.I.: Greedy attack and
  gumbel attack: Generating adversarial examples for discrete data. Journal of
  Machine Learning Research (43),  1--36 (2020)

\bibitem{yasunaga2017robust}
Yasunaga, M., Kasai, J., Radev, D.: Robust multilingual part-of-speech tagging
  via adversarial training. In: Proc. of NAACL-HLT. pp. 976--986 (2018)

\bibitem{zhai2008statistical}
Zhai, C.: Statistical language models for information retrieval. In:
  Proceedings of the Human Language Technology Conference of the {NAACL},
  Companion Volume: Tutorial Abstracts. pp.~3--4 (2007)

\bibitem{zhao2019improving}
Zhao, W., Wang, L., Shen, K., Jia, R., Liu, J.: Improving grammatical error
  correction via pre-training a copy-augmented architecture with unlabeled
  data. In: Proc. of NAACL-HLT. pp. 156--165 (2019)

\bibitem{zhao2017generating}
Zhao, Z., Dua, D., Singh, S.: Generating natural adversarial examples. In:
  Proc. of ICLR (2018)

\bibitem{zheng2016improving}
Zheng, S., Song, Y., Leung, T., Goodfellow, I.J.: Improving the robustness of
  deep neural networks via stability training. In: 2016 {IEEE} Conference on
  Computer Vision and Pattern Recognition, {CVPR} 2016, Las Vegas, NV, USA,
  June 27-30, 2016. pp. 4480--4488 (2016)

\end{thebibliography}

%
%
%
%





\end{document}